\pdfoutput=1

\documentclass[11pt]{article}

\usepackage[final]{acl}

\usepackage{times}
\usepackage{latexsym}

\usepackage[T1]{fontenc}
\usepackage{makecell}

\usepackage[utf8]{inputenc}

\usepackage{graphicx}
\graphicspath{ {./images/} }
\usepackage{enumitem}
\usepackage{hyperref}
\usepackage{booktabs}
\usepackage{subcaption} 
\usepackage{longtable} 
\usepackage{array}

\title{
Team Unibuc - NLP at GenAI Detection Task 1: \\
\textit{Qwen} it detect machine-generated text?
}

\author{
Teodor-George Marchitan$^{1, 3}$,
Claudiu Creanga$^{2, 3}$, Liviu P. Dinu$^{1, 3}$\\ 
  $^1$ Faculty of Mathematics and Computer Science, \\
  $^2$ Interdisciplinary School of Doctoral Studies, \\
  $^3$ HLT Research Center, \\
  University of Bucharest, Romania\\
  \small
{\tt teodor.marchitan@s.unibuc.ro, claudiu.creanga@s.unibuc.ro, ldinu@fmi.unibuc.ro}  \\
}

\vspace{2mm}
  
\begin{document}

\maketitle

\begin{abstract}

This paper describes the approach of the Unibuc - NLP team in tackling the Coling 2025 GenAI Workshop, Task 1: Binary Multilingual Machine-Generated Text Detection. We explored both masked language models and causal models. For Subtask A, our best model achieved \textbf{first-place} out of $36$ teams when looking at F1 Micro (Auxiliary Score) of $0.8333$, and \textbf{second-place} when looking at F1 Macro (Main Score) of $0.8301$. 

\end{abstract}

\section{Introduction}
Task 1 from the GenAI Content Detection Workshop \cite{wang2025genai} focuses on discerning whether a text sample is machine-generated or human-authored. With human ability to distinguish AI-generated text from human content near random chance, advanced automated systems are needed to ensure information integrity. Such systems are crucial for verifying content sources and countering unethical AI use, including propaganda, misinformation, deepfakes, and social manipulation, which pose significant societal risks.

The system developed for Task 1 subtask A is based on an LLM model where only the last layer and the classification head were trained for the downstream task, on the other hand, the system developed for subtask B is based on a transformer model with a classification head on top and it was completely fine-tuned using Low-Rank Adaptation (LoRA) \cite{hu2022lora}.

We made our models publicly available in a \href{https://github.com/ClaudiuCreanga/coling-2025-task-1}{GitHub Repository}.

\section{Background}

The competition had 3 tasks:
\begin{enumerate}
    \item Binary Multilingual Machine-Generated Text Detection (Human vs. Machine) with 2 SubTasks: English and Multilingual;
    \item AI vs. Human – Academic Essay Authenticity Challenge
    \item Cross-domain Machine-Generated Text Detection, which is the same challenge as Task 1, but the texts come from 8 domains. 
\end{enumerate}

We participated in Task 1 and achieved the top position in the Monolingual Subtask based on F1 Micro score, and secured the second position when considering F1 Macro.

\subsection{Dataset}

The data for this task is an extension the SemEval 2024 Task 8, which itself is based on the M4 dataset \cite{wang2024m4, wang2024mg-bench}. This dataset has many more examples, models and sources than the previous ones (\autoref{table:dataset_sizes}). 

\begin{table}
\begin{tabular}{l|c|c|c}
\hline
    \thead{Subtask} & \thead{Train} & \thead{Dev} & \thead{Test}  \\  \hline
    \texttt{A (mono)} & $610,767$ & $261,758$ & $73,941$ \\
    \texttt{B (multi)   } & $674,083$ & $288,894$ & $151,425$  \\
\hline
\end{tabular}
  \caption{Datasets sizes used in the Task 1 for each subtask.}
  \label{table:dataset_sizes}
\end{table}

\begin{figure}[htbp]
    \centering
    \includegraphics[width=0.9\linewidth]{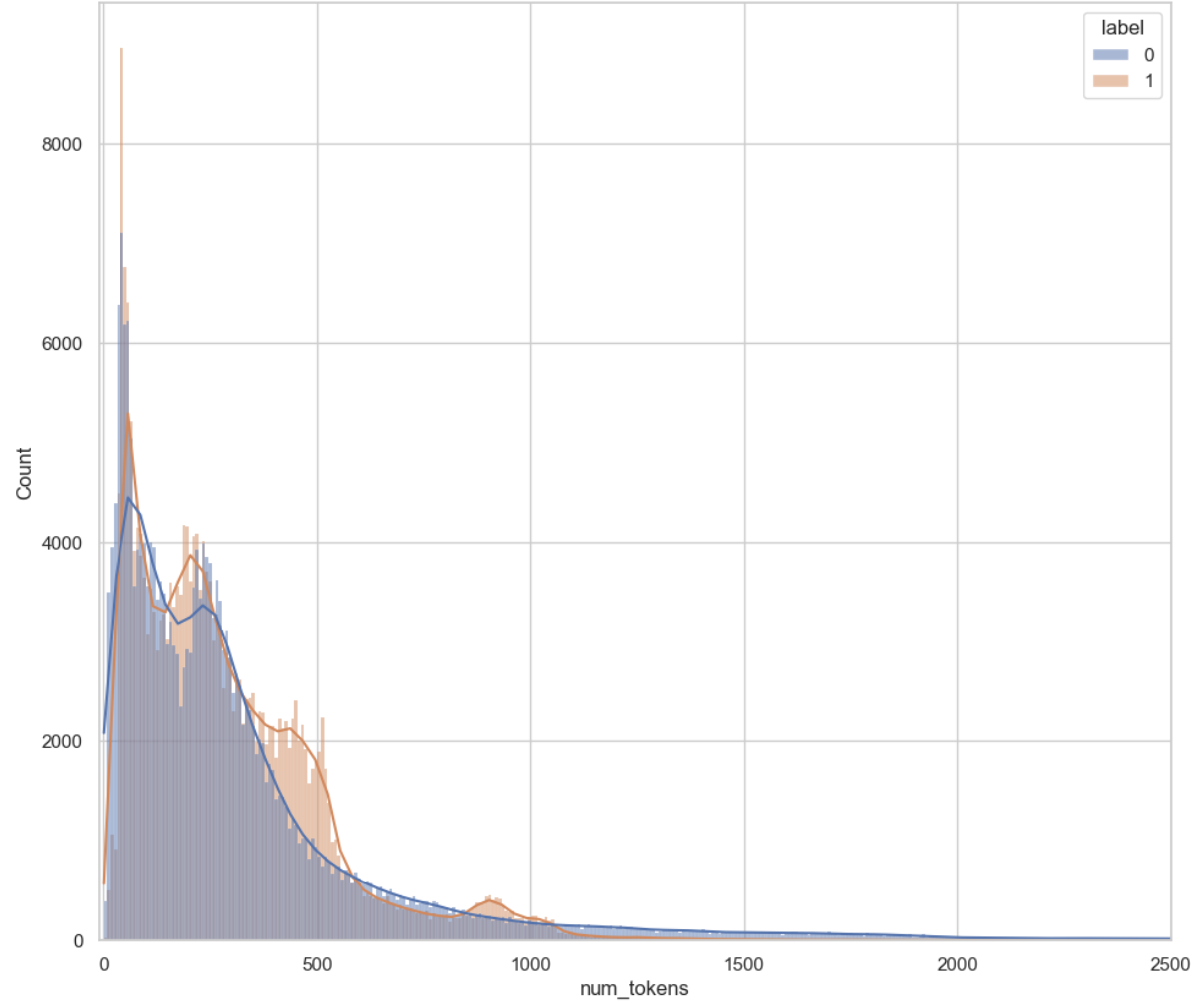}
    \caption{Subtask A: Distribution of token length for the training dataset.}
    \label{fig:tokens_train}
\end{figure}

\begin{figure}[htbp]
    \centering
    \includegraphics[width=0.9\linewidth]{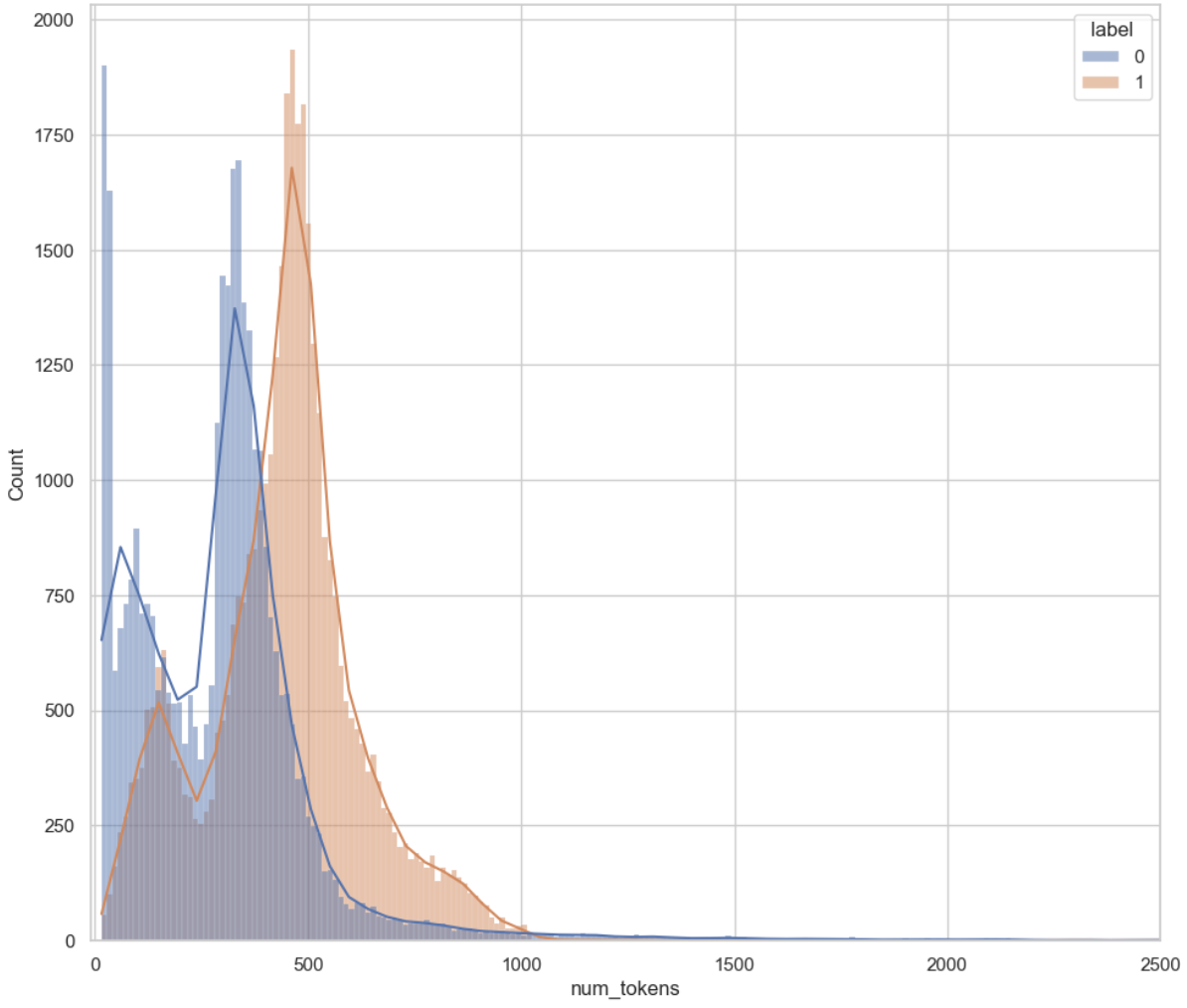}
    \caption{Subtask A: Distribution of token length for the test dataset. We can see it is significantly different from the training set. }
    \label{fig:tokens_test}
\end{figure}

Examining token count distribution reveals that, on average, the generated class in the test set has more tokens than in the training set (\autoref{fig:tokens_train} vs. \autoref{fig:tokens_test}).

\subsection{Previous Work}

Recent years have witnessed a significant evolution in language model capabilities, with models like GPT-2, GPT-3 and GPT-4 pushing the boundaries of machine-generated text. This advancement has made it increasingly challenging to distinguish between human-authored and machine-generated content. Early language models using top-k sampling often produced detectable patterns, like repetitive words, which machine learning models could exploit for identifying AI-generated text. However, advanced techniques like nucleus sampling have reduced these cues, making detection much harder \cite{ippolito-etal-2020-automatic}. 

While fine-tuning large language models for detection has shown some promise, as demonstrated by the success of RoBERTa in detecting GPT-2-generated text \cite{Solaiman2019ReleaseSA}, the increasing sophistication of these models continues to pose a significant challenge. Human evaluators, even for earlier models like GPT-2, struggled to accurately identify machine-generated content, achieving only around $70\%$ accuracy \cite{ippolito-etal-2020-automatic}. For more advanced models like GPT-3, human evaluators perform at chance levels, highlighting the limitations of human judgment in this domain \cite{clark-etal-2021-thats}. Given the rapid advancement of language models, there is an urgent need for further research into automated detection methods. It remains an open question whether we can develop systems capable of keeping pace with the evolving capabilities of generative models.

\section{System overview}

In this paper, we focused our research on two different system architectures: \textbf{Causal models} (\ref{subsec:causal_models}) for subtask 1 and  \textbf{Masked models} (\ref{subsec:masked_models}) for subtask 2.

\subsection{Causal models}\label{subsec:causal_models}

We conducted experiments using several large language models (LLMs), exploring both small and large variants to identify the model that achieved the best performance on our task. Among the models tested: BLOOM-560M \cite{workshop2023bloom176bparameteropenaccessmultilingual}), Llama-3.2-1B \cite{grattafiori2024llama3herdmodels}, the highest-performing model was based on Qwen, precisely the model Qwen2.5-0.5B \cite{qwen2.5}. Initially, the model displayed a tendency to over-fit on the majority class (F1 Macro: $0.7783$ and F1 Micro: $0.7868$ on the final test set), leading us to down-sample the training set to achieve a balanced $50$-$50$ distribution between the two classes. This adjustment helped mitigate over-fitting and improved the model's generalization. With this model we achieved first place on the leader-board by F1 Micro ($0.8333$) score and second place by F1 Macro score ($0.8301$). We additionally experimented with Gemini 1.5 Flash. However, due to limited resources, we were unable to fine-tune the model. Consequently, its accuracy was poor, nearing random chance levels.

We set the maximum number of tokens to $2048$ (based on \autoref{fig:tokens_train}) and froze all layers excepting the last one and the classification head ending up with $14,914,176$ ($3.02\%$) trainable parameters. For training we used a learning rate of $0.0002$, a weight decay of $0.01$, a batch size of $32$, and trained for a maximum of three epochs. Throughout the training process, we closely monitored both training and validation losses to assess the model's learning progress and prevent over-fitting.

As shown in \autoref{table:training_epochs}, the model demonstrated effective learning from the first epoch onward. The training loss continued to decrease steadily, reflecting improved performance on the training data. However, by the third epoch, the validation loss had reached a plateau, suggesting that further training would not yield additional gains and could potentially lead to over-fitting. We therefore halted training after the third epoch.

\begin{table}
\begin{tabular}{l|c|c|c}
\hline
    \thead{Epoch} & \thead{Train Loss} & \thead{Valid Loss} & \thead{Macro F1}  \\  \hline
    \texttt{1} & $0.11$ & $0.11$ & $0.950$ \\
    \texttt{2} & $0.07$ & $0.10$ & $0.960$  \\
    \texttt{3} & $0.04$ & $0.10$ & $0.966$  \\
\hline
\end{tabular}
  \caption{Training and Validation Loss alongside Macro F1 score for the $3$ epochs.}
  \label{table:training_epochs}
\end{table}

\begin{figure}[htbp]
    \centering
    \includegraphics[width=0.9\linewidth]{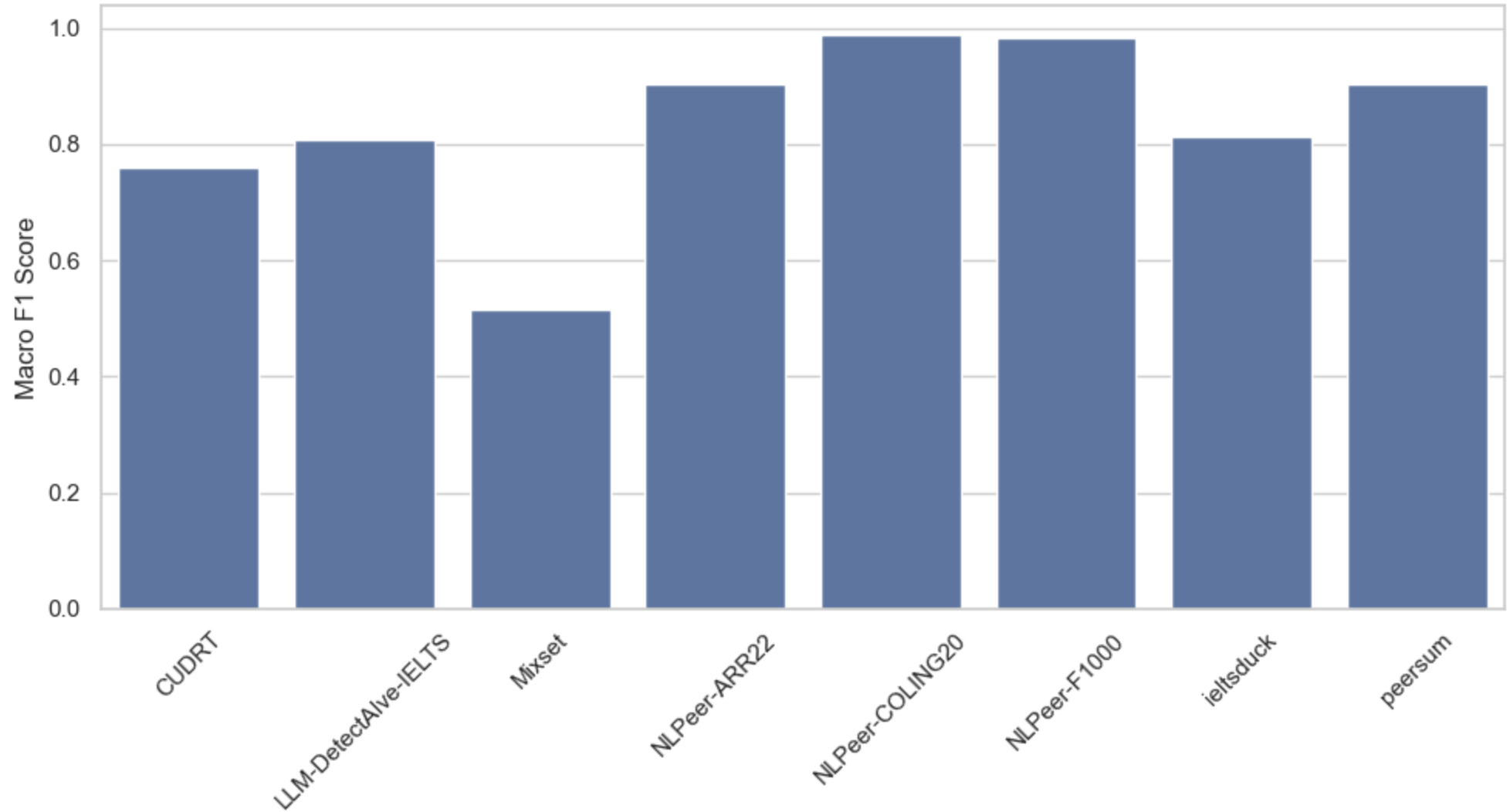}
    \caption{Subtask A: monolingual - accuracy by source for test set. We obtain best accuracy on NLPeer datasets, almost $100\%$. }
    \label{fig:accuracy_by_source}
\end{figure}

\subsection{Masked models}\label{subsec:masked_models}

The core of this architecture is based on transformer model XLM-Roberta-Base \cite{DBLP:journals/corr/abs-1911-02116} with a classification head on top consisting of 2 hidden layers with a dropout of $0.1$ between. We set the maximum number of tokens to $512$ and we truncated the longer text by keeping the first part of the text, as suggested in \cite{marchitan-etal}. We then fine-tuned the entire model using Low-Rank Adaptation (LoRA) \cite{hu2022lora} with the following hyperparameters: $r=4$, $lora\_alpha=8$, $lora\_dropout=0.25$ and we ended up with $739,586$ ($0.2653\%$) trainable parameters. The fine-tuning was done as in \cite{CREANGA20242100} for one epoch in batches of $16$ using the AdamW optimizer with a learning rate of $0.00005$, a weight decay of $0.002$ and warmup steps set to $10\%$ of the total number of sets (in our case $4213$). When fine-tuning this system, we have also used the class weights for the cross entropy function in order to make the model pay more attention to the minority class and penalize more the errors for this class.

\begin{figure*}[htbp]
    \centering
    \includegraphics[width=0.9\linewidth]{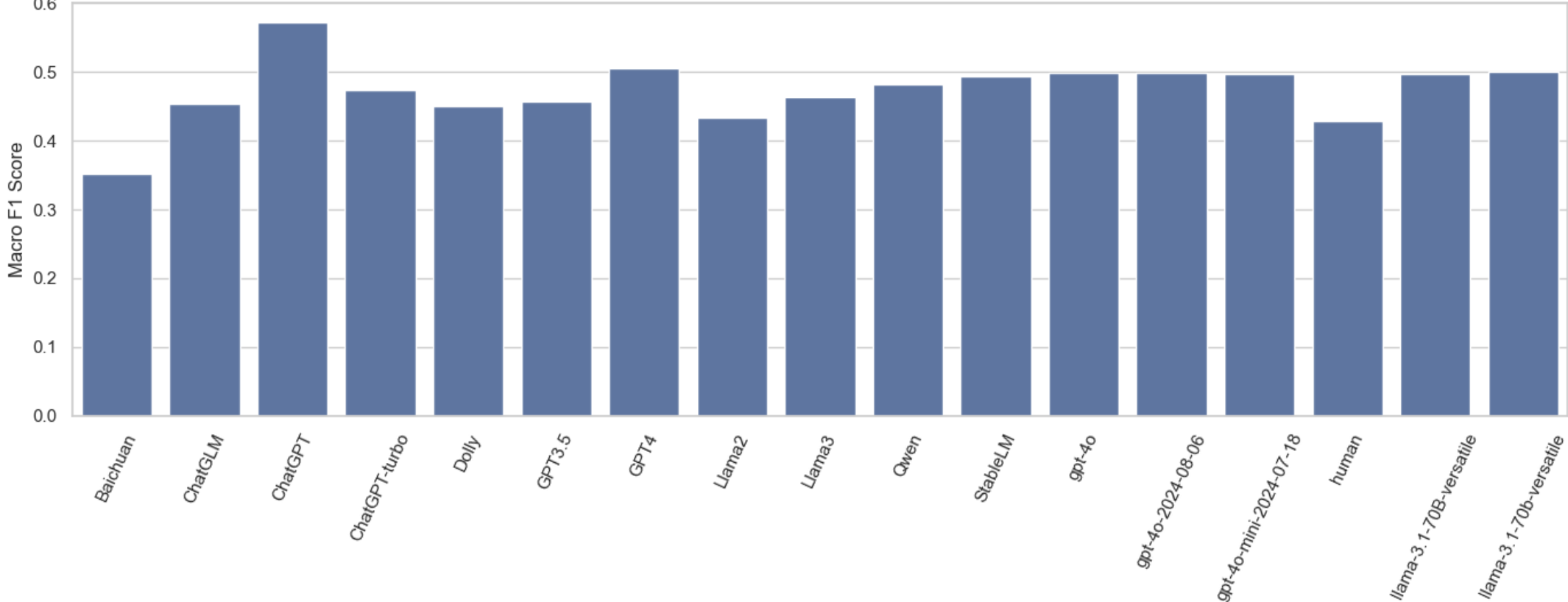}
    \caption{Subtask A: monolingual - accuracy by model for test set. We obtained best accuracy on ChatGPT, but otherwise there is not a lot of variation between models. }
    \label{fig:accuracy_by_model}
\end{figure*}

\section{Results}

We participated in Task 1 and achieved the \textbf{top position} in the Monolingual Subtask based on F1 Micro score, as shown in \autoref{table:team_results}, and secured the second position when considering F1 Macro. This reflects our model’s ability to consistently identify and classify instances correctly in the monolingual setting, achieving an F1 Micro score of $0.8333$. Our performance in F1 Macro, which captures how well our model handled imbalances across classes, placed us in a second position with a score of $0.8301$.

In the Multilingual Track, however, our model didn't do so well, securing 24th place with F1 scores of $0.66$ (Macro) and $0.67$ (Micro). This gap between monolingual and multilingual performance highlights the difficulties our model encountered when adapting to varied languages and possibly diverse linguistic structures in the multilingual setting.

\begin{table}
\begin{tabular}{l|c|c}
\hline
     & \thead{Score and Place \\ Track Monolingual}  & \thead{Score and Place \\ Track Multilingual} \\ \hline
    \texttt \textbf{F1 Macro} & $0.8301$ / \textbf{2}  & $0.66$ / $24$ \\
    \texttt \textbf{F1 Micro} & $0.8333$ / \textbf{1} & $0.67$ / $24$ \\
\hline
\end{tabular}
  \caption{Team Unibuc - NLP
  results on Task 1}
  \label{table:team_results}
\end{table}

\subsection{Error Analysis}

Examining the F1 Macro scores by model (\autoref{fig:accuracy_by_model}) reveals that our model achieves the highest accuracy on data generated by ChatGPT. This result may be influenced by the relatively small number of ChatGPT samples in the test set ($96$), which could make high performance on this subset more attainable. Notably, although ChatGPT data was not included in the training set, our model was able to generalize effectively to this unseen data, indicating strong generalization capabilities. In contrast, the model's lowest accuracy is on text generated by Baichuan, which, like ChatGPT, was also absent from the training set. The reduced accuracy on Baichuan text suggests that this style or structure might be more challenging for the model to handle. 

Analyzing the F1 Macro scores by source (\autoref{fig:accuracy_by_source}) reveals that our model achieves its highest accuracy on NLPeer-COLING20 and NLPeer-F1000 data, with scores approaching nearly $100\%$. This exceptional performance may be partly attributed to the limited sample sizes of these sources in the test set: NLPeer-COLING20 contains only $176$ samples, and NLPeer-F1000 has a medium sample size of $9,798$. Smaller sample sizes can lead to higher apparent accuracy due to reduced variance, which may amplify the model’s ability to fit well on these subsets. On the other hand, the model shows its lowest accuracy on the Mixset source, with an F1 Macro score around $0.5$. This significant drop suggests that the Mixset data presents more challenging language structures or varied writing styles that the model finds difficult to generalize.

\begin{figure}[htbp]
    \centering
    \includegraphics[width=0.9\linewidth]{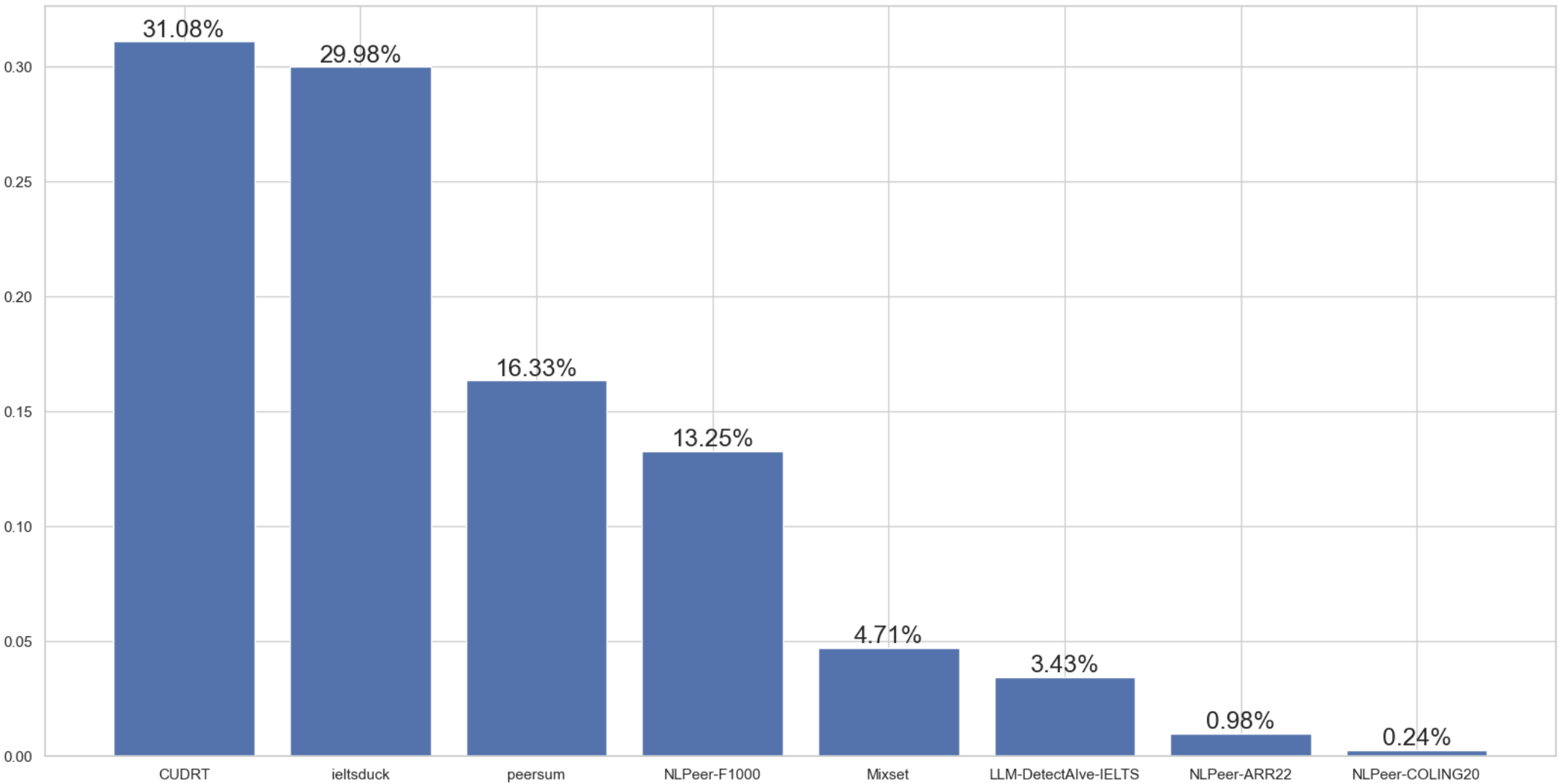}
    \caption{Subtask A: monolingual - Sources of the text in test dataset. In the training dataset we had only 3 sources: mage ($46\%$), m4gt ($42\%$) and hc3 ($11\%$).}
    \label{fig:test_by_source}
\end{figure}

Furthermore, as seen in the distribution of sources within the test set and training set (\autoref{fig:test_by_source}), none of these sources were present in the training data. Despite this, the model generalizes well to NLPeer sources, demonstrating its robustness in adapting to unseen data. 

\begin{figure}[htbp]
    \centering
    \includegraphics[width=0.9\linewidth]{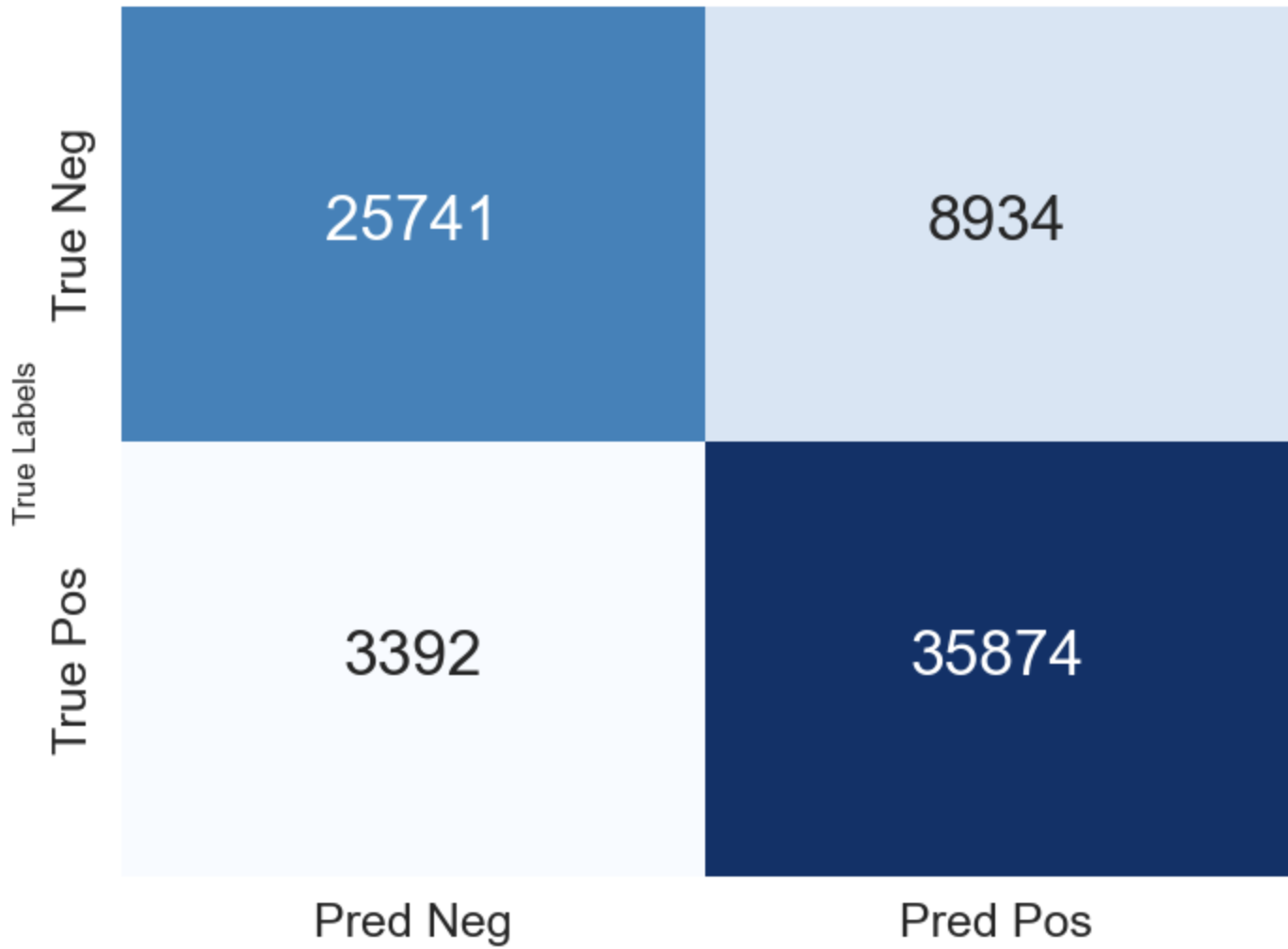}
    \caption{Subtask A: monolingual - Confusion matrix. Most predictions aligning along the correct diagonal: True negatives are $74\%$, False negatives are $25\%$, False positives are $8\%$, True positives are $91\%$.}
    \label{fig:confusion_matrix}
\end{figure}

Examining the confusion matrix (\autoref{fig:confusion_matrix}), we observe that the model achieves strong performance: $74\%$ of true negatives and $91\%$ of true positives are accurately classified. This indicates that the model is generally effective at distinguishing between classes. However, there is a tendency to over-predict the positive class. Specifically, the model made $44,808$ positive predictions compared to the $39,266$ actual positive examples in the dataset. This imbalance suggests that the model may be leaning towards identifying samples as positive, possibly due to certain linguistic patterns associated with the positive class. In contrast, the model under-predicts the negative class, suggesting that is necessary further fine-tuning to better capture the variations within human-generated text.

\section{Conclusions and Future Work}

In conclusion, our architecture and training methods produced good results for subtask A (securing the second place). However, our models demonstrated signs of over-fitting for subtask B.  Our future endeavors will explore several avenues: 
\begin{itemize}
    \item To improve our masked model's performance on the multilingual task, we will explore techniques such as language-specific fine-tuning, data augmentation, and regularization to prevent over-fitting.
    \item Utilize latent-space variables in our models and, with the help of high-level features such as event transitions or topic sequences, see if we can improve the accuracy and resilience of our model, especially under varied generation and adversarial settings.
\end{itemize}

\section*{Acknowledgements}

Research partially supported by the Ministry of Research, Innovation and Digitization, CNCS/CCCDI UEFISCDI,
SiRoLa project,  PN-IV-P1-PCE-2023-1701, within PNCDI IV, Romania.

\bibliography{anthology}

\appendix

\bigskip

\end{document}